\documentclass[11pt,a4paper]{article}
\usepackage[hyperref]{naaclhlt2019}
\usepackage{times}
\usepackage{latexsym}
\usepackage{amsmath}
\usepackage{amssymb}
\usepackage{graphicx}
\usepackage{rotating}
\usepackage{verbatim}
\usepackage{url}
\usepackage{subcaption} 

\aclfinalcopy 

\title{Inferring Which Medical Treatments Work from\\ Reports of Clinical Trials}

\author{Eric Lehman \\
  Northeastern University \\
  {\small\tt lehman.e@northeastern.edu} \\\And
  Jay B. DeYoung \\
  Northeastern University \\
  {\small\tt deyoung.j@northeastern.edu} \\\AND
  Regina Barzilay \\
  MIT \\
  {\small\tt regina@csail.mit.edu} \\\And
  Byron C. Wallace \\
  Northeastern University \\
  {\small\tt b.wallace@northeastern.edu} \\}

\date{}

\begin{document}
\maketitle
\begin{abstract}

How do we know if a particular medical treatment actually works?
Ideally one would consult all available evidence from relevant clinical trials.  Unfortunately, such results are primarily disseminated in natural language scientific articles, imposing substantial burden on those trying to make sense of them.
In this paper, we present a new task and corpus for making this unstructured evidence actionable.
The task entails inferring reported findings from a full-text article describing a randomized controlled trial (RCT) with respect to a given intervention, comparator, and outcome of interest, e.g., inferring if an article provides evidence supporting the use of \emph{aspirin} to reduce \emph{risk of stroke}, as compared to \emph{placebo}.

We present a new corpus for this task comprising 10,000+ prompts coupled with full-text articles describing RCTs.
Results using a suite of models --- ranging from heuristic (rule-based) approaches to attentive neural architectures --- demonstrate the difficulty of the task, which we believe largely owes to the lengthy, technical input texts. 
To facilitate further work on this important, challenging problem we make the corpus, documentation, a website and leaderboard, and code for baselines and evaluation available at {\small \url{http://evidence-inference.ebm-nlp.com/}}.

\end{abstract}

\section{Introduction}
\label{section:intro} 

Biomedical evidence is predominantly disseminated in unstructured, natural language scientific manuscripts that describe the conduct and results of randomized control trials (RCTs). 
The published evidence base is vast and expanding~\cite{bastian2010seventy}: at present more than 100 reports of RCTs are published every day, on average. It is thus time-consuming, and often practically impossible, to sort through all of the relevant published literature to robustly answer questions such as: \emph{Does infliximab reduce dysmenorrhea (pain) scores, relative to placebo}?

Given the critical role published reports of trials play in informing evidence-based care, organizations such as the Cochrane collaboration and groups at evidence-based practice centers (EPCs) are dedicated to manually synthesizing findings, but struggle to keep up with the literature \cite{tsafnat2013automation}. 
NLP can play a key role in automating this process, thereby mitigating costs and keeping treatment recommendations up-to-date with the evidence as it is published.  



\begin{figure}
\includegraphics[width=\linewidth]{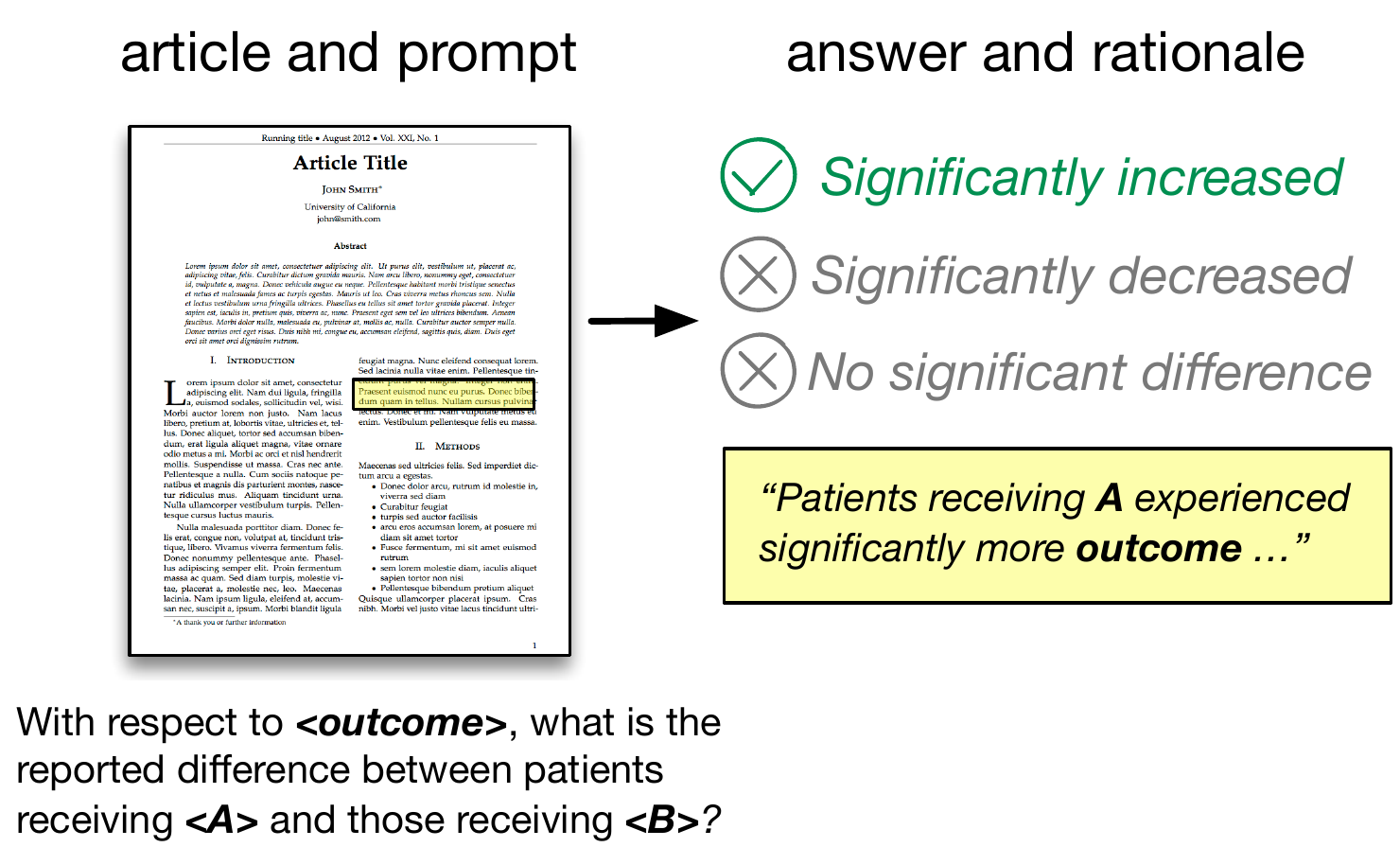} 
\caption{The task. Given a treatment {\bf A}, a comparator {\bf B}, and an {\bf outcome}, infer the reported relationship between {\bf A} and {\bf B} with respect to {\bf outcome}, and provide evidence supporting this from the text.} 
\label{figure:task}
\end{figure}


In this paper, we consider the task of inferring whether a given treatment is effective with respect to a specified outcome. Typically, this assessment is done relative to other treatment options (i.e., comparators). 
We assume the model is provided with a \emph{prompt} that specifies an intervention, a comparator, and an outcome, along with a full-text article.
The model is then to infer the reported findings with respect to this prompt (Figure \ref{figure:task}). 
From a healthcare perspective, this inference task is an essential step for automating extraction of actionable evidence from trial reports.


From an NLP standpoint, the proposed task can be seen as an instance of natural language inference~\cite{bowman-EtAl:2015:EMNLP}, viewing the article and prompt as the premise and hypothesis, respectively. However, the problem differs in a few important ways from existing NLP formulations. First, the inputs: prompts are brief ($\sim$13.5 words on average), but articles are long ($\sim$4200 words). Further, only a few snippets of the article will be relevant to the label for a given prompt. Second, prompts in this domain are structured, and include only a few types of key information: interventions, comparators, and outcomes. Methods that exploit this regularity are likely to be more accurate than generic inference algorithms.

Another interesting property of this task is that the target for an article depends on the interventions and outcome specified by a given prompt. Most articles report results for multiple interventions and outcomes: 67\% of articles in our corpus are associated with two or more prompts that have \emph{different} labels, e.g., indicating that a specific treatment was comparatively effective for one outcome but not for another. 
As a concrete example from our corpus, \emph{infliximab} was reported as realizing \emph{no significant difference} with respect to \emph{dysmenorrhea}, compared to a \emph{placebo}. 
But \emph{infliximab} was associated with a \emph{significant increase} in \emph{pain killer intake}, again compared to \emph{placebo}. 
Generally positive words in an article (e.g., ``improved") will confuse inference models that fail to account for this.
One may view these as built-in ``adversarial" examples \cite{jia-liang:2016:P16-1} for the task.

A key sub-problem is thus identifying snippet(s) of evidence in an article \emph{relevant to a given input prompt}. Attention mechanisms \cite{bahdanau2014neural} conditioned on prompts would seem a natural means to achieve this, and we do find that these achieve predictive gains, but they are modest. Existing attention variants seem to struggle to consistently attend to relevant evidence, even when explicitly pretrained using marked rationales. This corpus can facilitate further research in attention variants designed for lengthy inputs \cite{choi2017coarse,yang2016hierarchical}.


In sum, our contributions are threefold. We: (1) formulate a novel task (\emph{evidence inference}) that is both practically important and technically challenging; (2) Provide a new publicly-available corpus comprising 10,000+ evidence ``prompts", answers, supporting evidence spans, and associated full-text articles (\url{http://evidence-inference.ebm-nlp.com}) all manually annotated by medical doctors; (3) Develop baseline algorithms to establish state-of-the-art performance and highlight modeling challenges posed by this new task.

\section{Annotation}
\label{section:annotation} 
\begin{figure*}
\centering
\includegraphics[width=.85\textwidth]{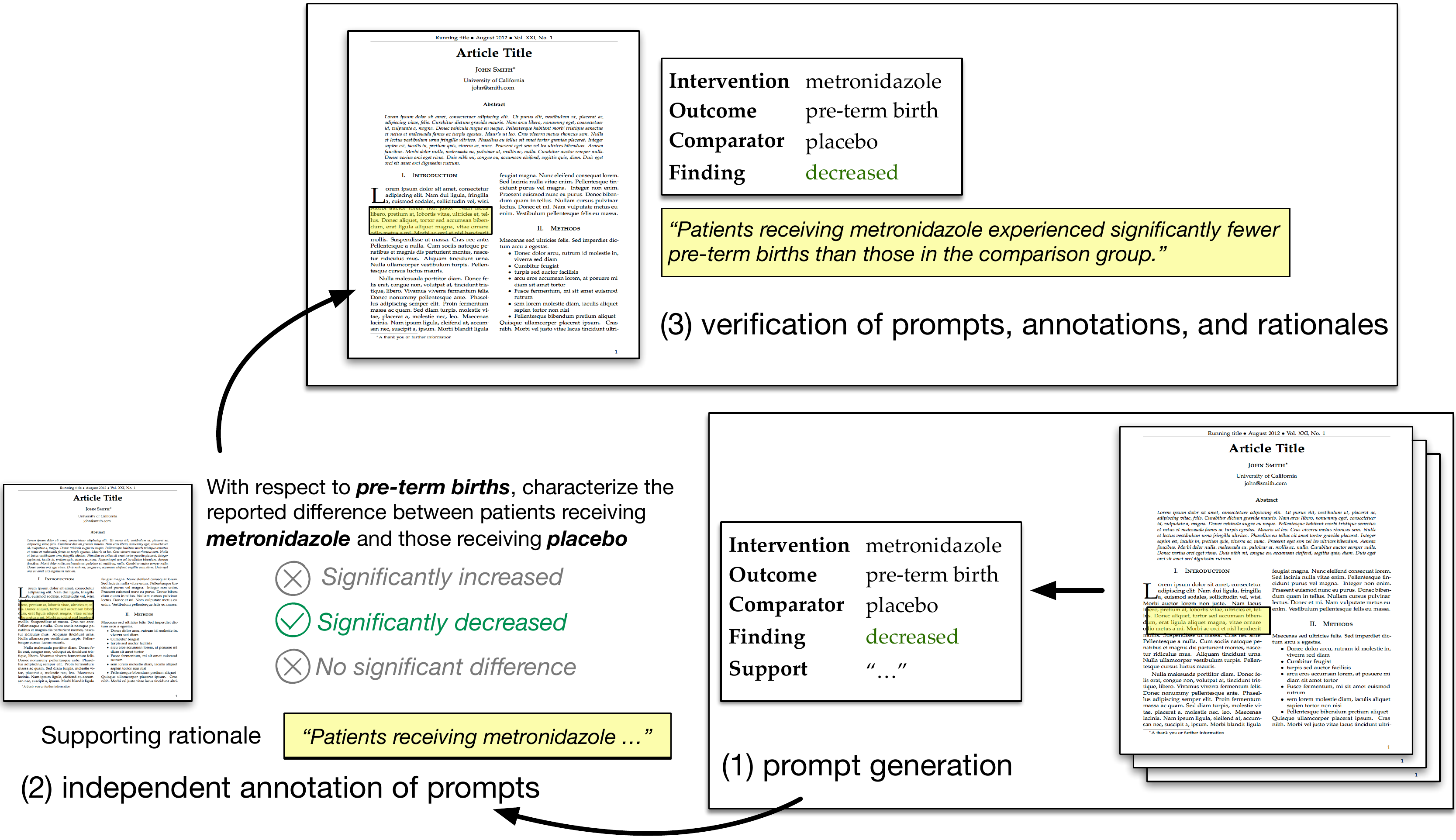}
\vspace{-.5em}
\caption{Schematic of the annotation process, performed by qualified (MD) annotators for all articles and prompts.}
\label{figure:annotation}
\end{figure*}

The specialized nature of this task necessitates adequate domain knowledge.
We thus recruited medical doctors (MDs) via the Upwork platform to perform annotation. 
Annotators were assigned to one of three mutually exclusive groups, responsible for: (1) prompt generation, (2) prompt and article annotation, and (3) verification. 
Figure \ref{figure:annotation} depicts the annotation process schematically; we describe these steps in more detail below. 

It is important to note that annotation was performed on full-texts, not just abstracts. 
Evidence relevant to a particular clinical question is quite often only available in the full text.
Indeed, in our dataset, the relevant evidence span was marked in the abstract only 40.5\% of the time.




\subsection{Prompt Generation}

This first task entails generating questions (or ``prompts'') that are answerable on the basis of a corresponding full-text article describing an RCT. 
Such prompts concern the comparison of specific interventions with respect to a particular outcome. 
Specifically, these questions ask whether an article reports that the specified intervention was found (in the described trial) to be significantly more effective than a comparator treatment, with respect to the outcome of interest. 

Prompt creators were instructed to identify a snippet, in a given full-text article, that reports a relationship between an intervention, comparator, and outcome. 
Generators were also asked to provide answers and accompanying rationales to the prompts that they provided; such supporting evidence is important for this task and domain.

As a concrete example, an example generated prompt for a trial described in \cite{marre2009liraglutide} specifies \emph{Proinsulin : insulin ratio} as the outcome of interest, \emph{liraglutide (1.8 mg) plus glimepiride} as the intervention, and \emph{rosiglitazone plus glimepiride} as the comparator. 
Liraglutide and rosiglitazone are both drugs that can be used to treat type 2 diabetes. 
In this case, use of the intervention (liraglutide) was reported to \emph{significantly decrease} the proinsulin to insulin ratio, as supported by the following evidence snippet extracted by the prompt creator: ``Reductions in the proinsulin : insulin ratio were greater with both liraglutide 1.2 and 1.8 mg compared with either rosiglitazone or placebo." 

Trials typically report results for multiple outcomes, and often for more than two interventions. 
As discussed above, results for these will often differ. 
For instance, \emph{postprandial plasma glucose} was another outcome reported in the aforementioned trial report, and \emph{placebo plus glimepeiride} was considered as another comparator.  
Therefore, we instructed prompt generators to create multiple prompts for each full-text article. 
On average, this yielded 4.19 distinct prompts per article.\footnote{We restricted generators to creating at most five prompts for a given article; prior to imposing this constraint, annotators would sometimes generate $>$10 prompts per article.}



Articles may be deemed \emph{invalid} for a few reasons, chiefly for not describing RCTs.\footnote{We used the RobotReviewer RCT classifier, which improves upon the standard MEDLINE RCT filter \cite{marshall2018machine}, but some false positives remain.} Of 3525 articles considered, 1106 were marked invalid (31.4\%). The prompt generators provided valid answers and rationales in 95.9\% and 97.8\% of cases, respectively, as per the verifier.



To summarize: prompt creation entails specifying answerable clinical questions, along with answers to these and supporting rationales (evidence snippets from the text).
This task is the most laborious step in the annotation process.

\subsection{Prompt Annotation}

For this task, annotators were asked to answer prompts on the basis of a particular article. 
More specifically, given an evidence prompt articulating an intervention, comparator, and outcome (generated as described above), the task was to determine whether the associated article reports results indicating that the intervention \emph{significantly increased}, \emph{significantly decreased}, or realized \emph{no significant difference}, relative to the comparator and with respect to the outcome.
The annotator was also asked to mark a snippet of text supporting their response.
Annotators also had the option to mark prompts as \emph{invalid}, e.g., if the prompt did not seem answerable on the basis of the article.  


Annotations collected in this step are redundant with the classification and rationale independently provided by the prompt generator in the preceding step; this is by design to ensure robust, high-quality annotations.

\subsection{Verification}

The final task in our annotation process entails a worker verifying the prompts and responses generated in the previous two steps.
The verifier is here responsible for checking both whether the prompt (i.e., question) is valid and can be answered from the text, and whether the responses provided are accurate. 
Verifiers also assess whether the associated supporting evidence provided is reasonable. 

Verification is a relatively easy task, because the verifier is directly provided all information relevant to making a quality judgment. 
Nonetheless, this step decidedly improved data quality: 3.8\% of prompts, 6.7\% of answers, and 7.1\% of rationales (supporting evidence snippets) were marked as invalid.
All invalid prompts were removed from the corpus; so too were all prompts for which the verifier rejected all answers or all rationales. 


\subsection{Task Refinement}
In an initial pilot round, we acquired annotations on 10 articles, yielding 93 prompts. 
Three medical doctors (MDs) were tasked with answering these prompts, achieving an agreement of 0.58 (Krippendorf's $\alpha$).
To improve this poor agreement, we provided personalized feedback that addressed systematic issues we observed. 
Following this feedback, the MDs were asked to re-examine the same set of prompts and update their responses if they felt it appropriate to do so.
This resulted in a much improved agreement of $\alpha$=0.84.

To verify that this agreement held beyond the specific set with respect to which we provided feedback, we subsequently assigned an additional 113 prompts to the annotators. 
As measured over these 113 prompts, the three annotators exhibited relatively high agreement between themselves and with the prompt generator (Krippendorf's $\alpha$ of 0.75 and 0.80, respectively). 





\section{Dataset Statistics}


\begin{table*}
\small
\centering
\begin{tabular}{llll|l}
\hline
                          &   Train                &  Dev     &    Test      &  Total          \\
\hline
 Number of prompts        & 8168               & 1004            & 965             & 10137              \\
 Number of articles       & 1931               & 248             & 240             & 2419               \\
 Label counts (-1 / 0 / 1) & 1981 / 3619 / 2568 & 232 / 448 / 324 & 215 / 403 / 347 & 2428 / 4470 / 3239 \\
\hline
\end{tabular}
\caption{Corpus statistics. Labels -1, 0, 1 indicate \emph{significantly decreased}, \emph{no significant difference} and \emph{significantly increased}, respectively.}
\vspace{-1em}
\label{table:counts}
\end{table*}

We hired 16 doctors from Upwork and split them at random into groups: 10 for prompt generation, 3 for annotation, and 3 for verification.\footnote{One doctor from verification was moved to annotation in order to increase productivity.} 
In total, we have acquired 10,137 annotated prompts for 2,419 unique articles. 
For each of these prompts, we have at least two independent sets of labels and associated rationales (supporting snippets). 



We additionally calculated agreement between prompt generators, annotators and verifiers using Krippendorf's $\alpha$.
To calculate this, we converted the verifier's binary labels of valid or not to the label with which they agreed. 
This yields $\alpha=0.88$.
Removing the verifier annotations from the calculation results in $\alpha$ = 0.86.

Intervention, outcome, and comparator strings contain on average 5.1, 5.3, and 3.4 tokens, respectively. Articles comprise a mean of 4.2k tokens. 

We provide additional details concerning the dataset in the Appendix. 

\section{Models}

We experimented with a suite of models to establish performance on this task, which we explain below in increasing order of complexity.

\begin{figure}
\centering
\includegraphics[width=\linewidth]{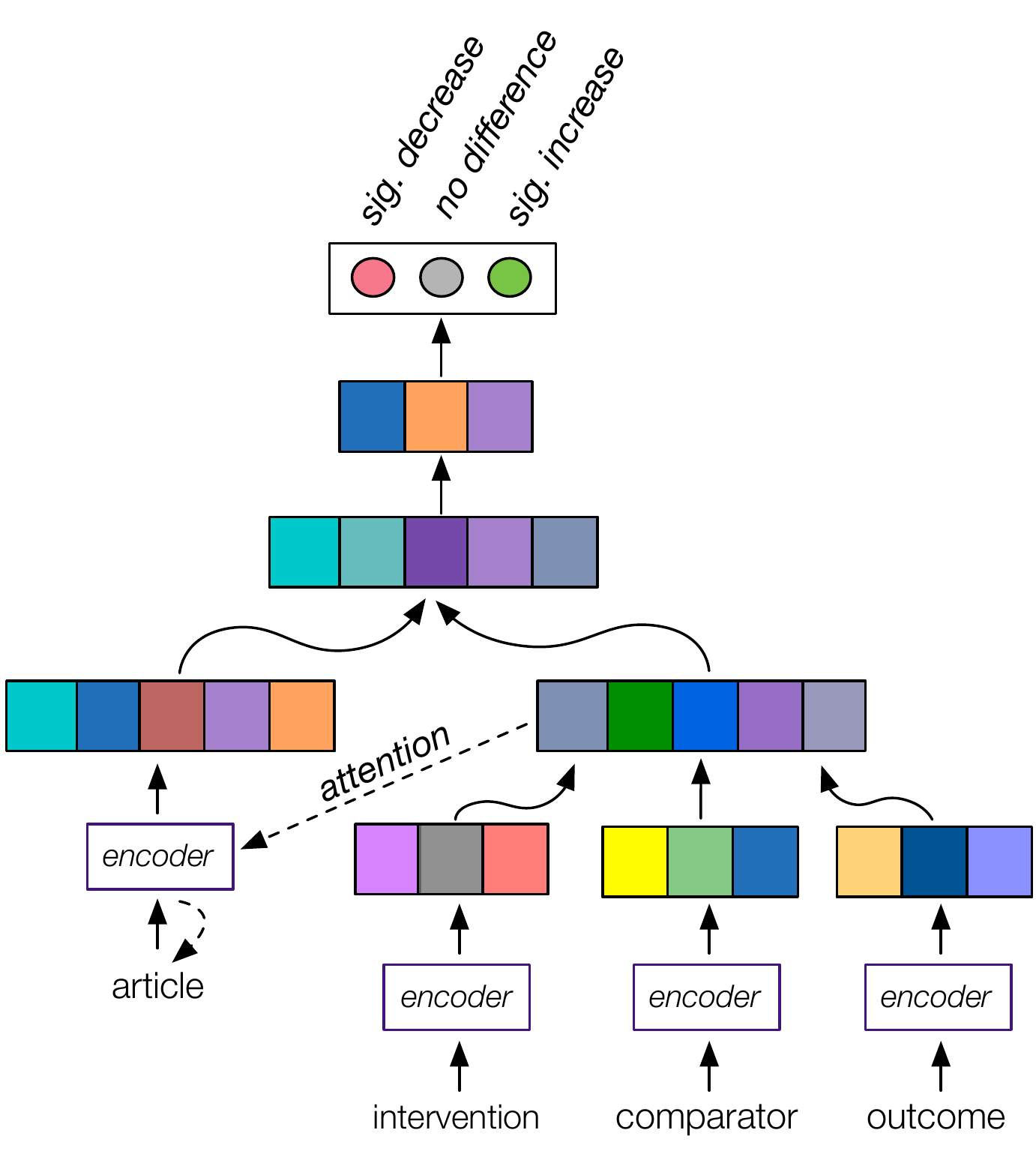}
\caption{A generalized illustration of the neural network we propose for evidence inference. }
\label{figure:model0}
\vspace{-1em}
\end{figure}

\subsection{Baselines}

\noindent {\bf Majority}. Predict the most common class, i.e., \emph{no significant difference}. 

\vspace{.5em}

\noindent {\bf Heuristics}. This entails two parts: (1) finding the sentence that contains the answer, and (2) interpreting the sentence that possesses the evidence. The first step of this process is achieved through locating the sentence that has the most overlap with words in the outcome, intervention, and comparator. Afterwards, we search for reported $p$-values within the identified sentence, and evaluate whether they seem significant. We provide a detailed description of the heuristics model in Section A of the Appendix.

\vspace{.5em}

\noindent {\bf Logistic Regression}. A standard logistic regression model trained on top of binary bag-of-words representations of articles and intervention, comprator and outcome (ICO) frames --- these are concatenated to form inputs. We use a vocabulary size of 20k (based on frequency of occurrence), thus yielding an input size of 80k. 

\vspace{.5em}

\noindent {\bf Neural Network Variants}. 
We encode the intervention, comparator, and outcome strings accompanying a prompt into vectors $i$, $c$, and $o$, respectively.
Similarly, we encode the article itself into a vector $a$.
We experimented with several encoder options, including simple bag-of-words style encoding (i.e., averaging constituent word vectors) and RNNs. For the latter, we specifically pass a Gated Recurrent Unit \cite{cho2014learning}, or GRU, over inputs, yielding hidden states for each article token.

In preliminary experiments we found that simple averages over token embeddings worked well for encoding prompts ($i$, $c$ and $o$), likely because they tend to be quite short.
But this encoding works terribly for articles, due to their length. 
Therefore, we use a GRU to encode articles (uni-directional, as bi-directional added complexity without improving results). 

In the simplest neural model variant, we simply concatenate the encoded article and ICO frame into a vector $[a; i; c; o]$ which is then passed through a feedforward network with a single hidden layer to allow interactions between the prompt and article text.\footnote{We use a linear hidden layer; experiments adding a nonlinearity (ReLU) did not affect results.} As discussed in detail below, we experiment with a variety of attention mechanisms imposed over article tokens.

\subsection{Finding the Evidence}
\label{section:identifying-evidence}
Exploiting the spans of evidence marked as supporting assessments should improve the predictive performance of models. 
An additional advantage of modeling this explicitly is that models will then be able to provide \emph{rationales} for decisions \cite{lei2016rationalizing,zheng:2016:EMNLP,zaidan2007using}, i.e., snippets of text that support predictions.
We therefore experiment with model variants that classify input tokens as being relevant evidence (or not) prior to performing inference. 

We consider both \emph{pipeline} and \emph{joint} instantiations of such models. 
In the former type, the model first identifies spans in the text and then passes these forward to an independent component that makes predictions on the basis of these. 
In models of the latter type, evidence span tagging and document-level inference is performed end-to-end. 
Evidence snippets are not restricted to sentence boundaries (i.e., are token-wise), 
but we also consider model variants that relax evidence span tagging to a sentence labeling task (classifying sentences as containing \emph{any} evidence tokens, or not).
In either case, which spans are relevant will depend on the prompt assessed. 
Thus, we consider and contrast variants that condition evidence span prediction on the input prompt.

We consider both linear and neural models. 
For the former, we train two logistic regression models over bag-of-words input representations.
The first predicts whether or not a given sentence contains any evidence tokens. 
Document predictions are then made via a second (independent) logistic regression model that consumes aggregate bag-of-words representations of only those sentences predicted to contain evidence. 
This is a linear model, and thus does not accommodate interactions; we therefore consider only an unconditioned version.

For our pipeline neural model, we first induce vector representations of article sentences via a GRU, and these are then passed through a binary classification layer. 
To allow interactions between the input prompt and the sentence being classified, we also consider a conditioned variant in which the sentence classification model is provided the induced vector representations of the prompt elements alongside the sentence vector. 



For end-to-end models, we capitalize on attention mechanisms that learn to focus on (contextualized hidden representations of) individual article tokens prior to making a prediction.
We consider several variants of attention, and we explore directly pretraining these using the marked evidence spans available in the training data. 

The simplest attention module we consider is unconditioned; we simply learn weights $W$ that scores hidden states $h_a$ output from the article encoder. Concretely,
\begin{equation}
    \alpha = {\text{softmax}}\{w_\alpha \cdot H_a\}
    \label{eq:vanilla-attn}
\end{equation}

\noindent where $w_\alpha \in \mathcal{R}^{1 \times d}$ and $H_a \in \mathcal{R}^{d \times |a|}$, denoting hidden size by $d$ and article length by $|a|$.\footnote{We have elided bias terms for presentation. }

\begin{figure}
\includegraphics[width=\linewidth]{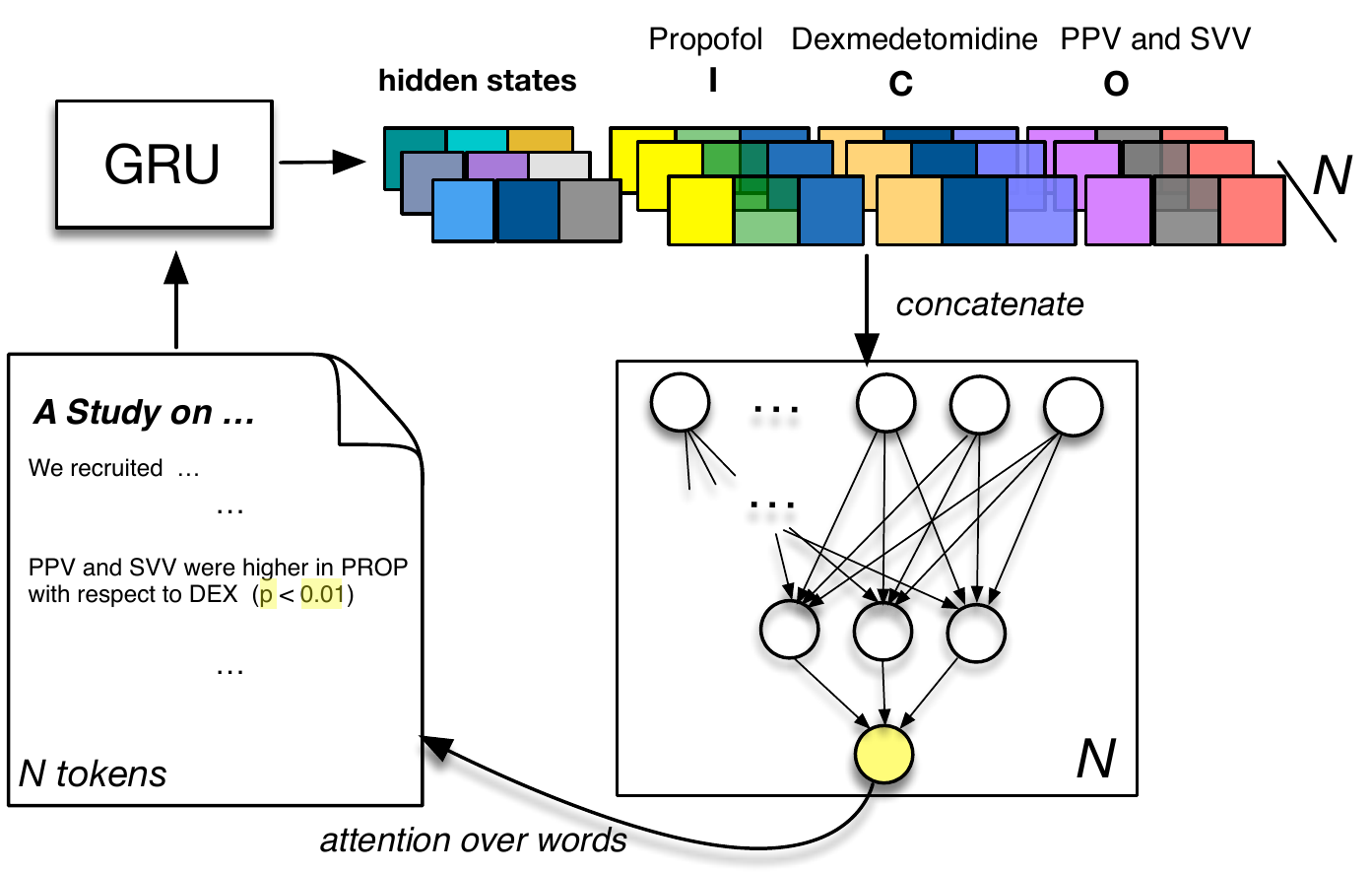}
\caption{The proposed conditional attention variant. ICO frame embeddings are concatenated to hidden state vectors from the GRU and fed through an MLP to induce attention weights.}
\vspace{-1em}
\label{figure:cond-attn}
\end{figure}

However, the text span relevant to a classification will depend on the prompt under consideration. 
We thus also consider a conditioned variant of attention. In this version we concatenate the $i$, $c$, and $o$ vectors induced by our encoders to the hidden states. Abusing notation a bit, denote the matrix in which we concatenate the $i$, $c$, and $o$ vectors to each column in $H_a$ by $[H_a; i; c; o]$. 
We then consider an attention variant that passes this concatenated representation through a single hidden layer to score tokens (Figure \ref{figure:cond-attn}). 
\begin{equation}
    \alpha = {\text{softmax}}\{v^t_\alpha \cdot {\text{tanh}}(W_\alpha \cdot [H_a; i; c; o])\}
    \label{eq:conditioned-attn}
\end{equation}


We consider two ways of converting the provided evidence spans into targets: (1) Imposing a uniform distribution over marked evidence tokens; (2) Setting the target for all marked evidence tokens as 1. 
In both cases we treat the absence of annotations on a token as an implicit negative target (0).
It is important to note that the model will see the same article multiple times during training with different evidence span targets, one for each prompt in the train set. 
The snippet of text that supports a particular assessment naturally depends on the prompt under consideration. 
Unconditioned attention variants will thus, by construction, be unable to attend exclusively to the relevant spans of text for across all prompts.

When training with binary targets, we consider two specifications: one in which the outputs of the attention model are independent (per-token) sigmoids indicating whether or not a word belongs to an evidence span, and another in which attention weights are normalized via a softmax over tokens. 
The latter is standard, although per-token attention has been previously proposed \cite{kim2017structured}.

\section{Experimental Details}

\subsection{Development Setting}
\label{subsection:development-setting}

During model development, we used 90\% of the train set for training, and the remaining 10\% as to monitor performance over epochs. 
To iteratively assess and refine models during this development phase, we used the standardized validation set.
All decisions regarding final experiments to run were made using this validation set, prior to evaluating models on the held-out test set of articles. 
Results reported in this paper are on the final test set.
Note that we report averages for neural models (over five runs) to mitigate noise due to random initialization and fitting.

\subsection{Training Details and Hyperparameters}

All neural variants were trained up to 50 epochs with a patience of 10 epochs.
We monitored performance during training on a nested development set and retained the model that achieved the highest F1 score on this.
For GRU encoders we used 32 hidden units. 
All models for the primary task were trained with batch sizes of 32.

We initialized word embeddings to pretrained word vectors induced over a large set of PubMed abstracts \cite{moendistributional}. 
Given the modest training dataset size, we did not fine-tune these. 


\begin{table*}
\centering
\small 
\begin{tabular}{l|rrr|r}
\hline
 Model                                     &    Precision &   Recall &    F1 & Evidence token AUC / mass \\
\hline
 Majority                                  & 0.138 & 0.333 & 0.195 & --\\
 Heuristics                                & 0.431 & 0.389 & 0.354 & 0.682 / 0.025  \\
 \hline
 Logistic regression                        & 0.409 & 0.400 & 0.388 & -- \\ 
 Pipeline logistic regression              & 0.452 & 0.429 & 0.423 & 0.523 / 0.012 \\
 \hline 
 Pipeline neural network & 0.422 & 0.405 & 0.402 & 0.847 / 0.080  \\
 \hspace{.3em} + Conditioned & 0.426 & 0.420 & 0.417 & 0.863 / 0.062 \\
 End-to-end neural network       & 0.471  & 0.439 & 0.440  & --\\ 
  \hspace{.3em}   + Attention          & 0.528 &  0.507 & 0.508 &  0.759 / 0.047 \\
  \hspace{.3em}   + Pretrain attention$^\dagger$ & 0.527 & 0.507 & 0.505 & 0.880 / 0.129 \\ 
  \hspace{.3em}  + Conditional attention &  0.522  & 0.504   & 0.505 &  0.706 / 0.059 \\
   \hspace{.3em}  + Pretrain conditional attention$^\dagger$  & 0.531 & 0.519 & 0.520 & 0.836 / 0.125 \\ 
\hline
\end{tabular}
\caption{Summary results on the evidence inference task test set, averaged over five independent runs (with independent random initialization values). Metrics are macro-averages over classes. Evidence token AUC and mass (last column) quantify identification of relevant supporting tokens. Models in the top two rows perform no learning; the second two correspond to linear models; the rest are neural model variants. $\dagger$ indicates `token-wise` attention pretraining; we report results for alternative attention losses in the Appendix.} \vspace{-.5em}
\label{table:main-results}
\end{table*}

\subsection{Attention pretraining Details}
\label{section:attn-pretrain} 
We use the manually marked supporting snippets as explicit, intermediate supervision for pretraining the attention mechanisms described in \ref{section:identifying-evidence}. 
More specifically, we pretrain the attentional model components for both conditioned and unconditioned attention variants. 

Concretely, we minimize token-wise binary cross entropy loss with respect to one of the two token-wise targets delineated in the preceding Section. We normalize loss per batch by the number of constituent tokens, using batch sizes of 16.\footnote{Memory constraints precluded larger batches.}
We monitor token-wise AUC with respect to the reference evidence span annotations marked in the held-out validation set mention in Section \ref{subsection:development-setting}. 
We retained the model that achieved the best AUC measured over fifty epochs of attention pretraining (again with a patience of ten) and used these weights as initialization values for fine-tuning the end-to-end inference network.\footnote{We also experimented with `freezing' attention module parameters during fine-tuning, but this performed poorly.}

We trained all models with the Adam optimizer using the parameters suggested by \cite{KingmaB14}. 
We  trained using PyTorch \cite{paszke2017automatic}, {\tt v 1.0.1.post2}.\footnote{This is a nightly build, used due to a dependence on recently introduced RNN utilities.}
Code for our models and to reproduce our results is available at: \url{https://github.com/jayded/evidence-inference}.

\subsection{Pipeline Model Training Details}
Pipeline models first attempt to identify sentences containing evidence.
To train these, we generalize token-wise annotations to sentences such that a sentence is labeled 1 if it contains any evidence tokens, and 0 otherwise. 
We then trained the sentence tagging models described above with these labels, monitoring loss on a nested validation set and retaining the best observed model over 50 epochs.
The document-level model subsequently consumes only sentences tagged as relevant.

\section{Results}
\vspace{-.25em}

\begin{table}
\centering
\small 
\begin{tabular}{lrrr}
\hline
Class & Precision &   Recall &    F1 \\ 
\hline
Sig. decreased & 0.448 & 0.334 & 0.380 \\ 
No sig. diff.  & 0.586 & 0.636 & 0.610 \\
Sig. increased & 0.556 & 0.585 & 0.572 
\end{tabular}
\caption{Average per-class test performance of best overall model (\emph{pretrain conditional attention}).}
\label{table:per-class-best}
\vspace{-.75em}
\end{table}

\subsection{Main Task Results}

Results on the main task for proposed model variants are reported in Table \ref{table:main-results}. 
These are averages over five independent runs, to ensure relatively robust measures of model performance.\footnote{These models exhibit a fair amount of variance; we report ranges over the validation set in the Appendix.}
The best performing model exploits pretrained conditional attention.
For the leaderboard we assume a single set of model predictions. To generate these we evaluated models on the test set using the versions that realized the strongest observed performance on the validation set over the aforementioned five runs/initializations. 
The best performing model (and hence current leader) is the variant that uses pretrained, conditional attention, which aligns with the average results in Table \ref{table:main-results}.
Table \ref{table:leaderboard} reports the results here, along with a more standard attentive architecture for context.


\begin{table}
\centering
\small 
\begin{tabular}{lrrr}
\hline
Model & Precision &   Recall &  F1 \\ 
\hline
NN + Attention & 0.518 & 0.503 & 0.505 \\
{\bf NN + pretrain cond. attn.} & {\bf 0.533} & {\bf 0.530} & {\bf 0.531}
\end{tabular}
\vspace{-.75em}
\caption{Leaderboard results.}
\label{table:leaderboard}
\vspace{-1em}
\end{table}

To highlight the importance of identifying relevant spans to inform predictions, we present results achieved when these are provided directly to models via an `oracle' prior to prediction in Table \ref{table:cheating}.
Access to this oracle yields a 20+ point jump in F1, indicating that accurately extracting the relevant evidence is critical. 
Below (Section \ref{section:analyze-attn}) we attempt to elucidate how well (or poorly) attention mechanisms fail to find supporting evidence.


A natural question that arises in NLP tasks in which the output depends on both a document and a question (here, a prompt) is: how much does the latter in fact influence model predictions \cite{kaushik2018much}? We explore this in Table \ref{table:removing-things}.
Relying only on the prompt (ignoring the article completely) achieves surprisingly strong performance, outperforming a vanilla neural model (sans attention). 
This is not entirely unreasonable, as certain intervention types will tend to correlate with significant vs insignificant findings, i.e., the prompt itself contains signal.
The neural model without attention is likely simply unable to extract meaningful signal from lengthy articles, and so induced representations merely add noise.
By contrast, ignoring the prompt severely degrades performance.

\begin{table}
\centering
\small 
\begin{tabular}{lrrr}
\hline
 Model                                     &    Precision &   Recall &    F1 \\ 
\hline
Best NN & 0.531 & 0.519 & 0.520  \\
NN (no attention) & 0.471  & 0.439 & 0.440   \\
 - prompt &  0.344 & 0.340  & 0.324 \\
 - article &  0.489 & 0.468 & 0.472 \\
\end{tabular}
\vspace{-.5em}
\caption{Average results achieved (macro-averages over five runs) by the neural model when it is provided only the article or only the prompt. We reproduce results for the best model from Table \ref{table:main-results} and the vanilla (no attention) end-to-end neural network for context. }
\label{table:removing-things}
\vspace{-.5em}
\end{table}

\subsection{Analyzing Attention}
\label{section:analyze-attn}

\begin{table}
\centering
\small 
\begin{tabular}{lrrr}
\hline
 Model                                     &    Precision &   Recall &    F1 \\ 
\hline
 Heuristics                                & 0.492 & 0.457 & 0.453  \\
 Logistic regression                       & 0.732 & 0.734 & 0.731  \\
 Neural network &  0.740 & 0.739 & 0.739 \\
\end{tabular}
\caption{Average results achieved when models are provided directly with the reference evidence spans.}
\label{table:cheating}
\vspace{-.75em}
\end{table}

To provide a sense of how well models are able to identify relevant evidence (i.e., tokens in the supporting snippets marked by annotators), we report token AUCs and \emph{evidence masses} for all models that assign scores to words.
The former captures how well models discriminate evidence from non-evidence tokens in general; the latter measures the relative amount of attention payed to evidence tokens. 
Concretely we calculate attention mass as a sum of the normalized attention scores assigned to words that belong to reference evidence spans.
Thus, e.g., if the evidence token mass were 1, this would mean the model attended to \emph{only} relevant evidence, ignoring all other tokens.
We also experimented with optimizing for this directly during attention pretraining (see Appendix).

Aside from the \emph{Majority} and \emph{LR} baselines, all of the models explored generate scores encoding token relevance, either explicitly or implicitly.
Attentive neural variants induce these by scoring contextualized representations of tokens $t$, $h_t$ for relevance.
Pipelined models score sentences, not tokens. 
For comparison across models, we assign the probability predicted for a given sentence to all of the words that it contains.
Note that the maximum evidence token AUC achievable when selecting a sentence is $\sim$0.92.

Qualitatively, we observe that attention weights often, though not always, square with intuition. 
In a test example wherein the intervention is \emph{propofol}, the comparator \emph{dexmedetomidine}, and the outcome \emph{Stroke Volume Variation (SVV) and Pulse Pressure Variation (PPV)}, the conditioned, pretrained attention model focuses on tokens `svv', `versus' (suggesting comparison), and $p$-value indicators (`p', `01'). 
This is not surprising given the reasonably high evidence token AUC achieved by this model.

\begin{figure}
    \includegraphics[width=0.45\textwidth]{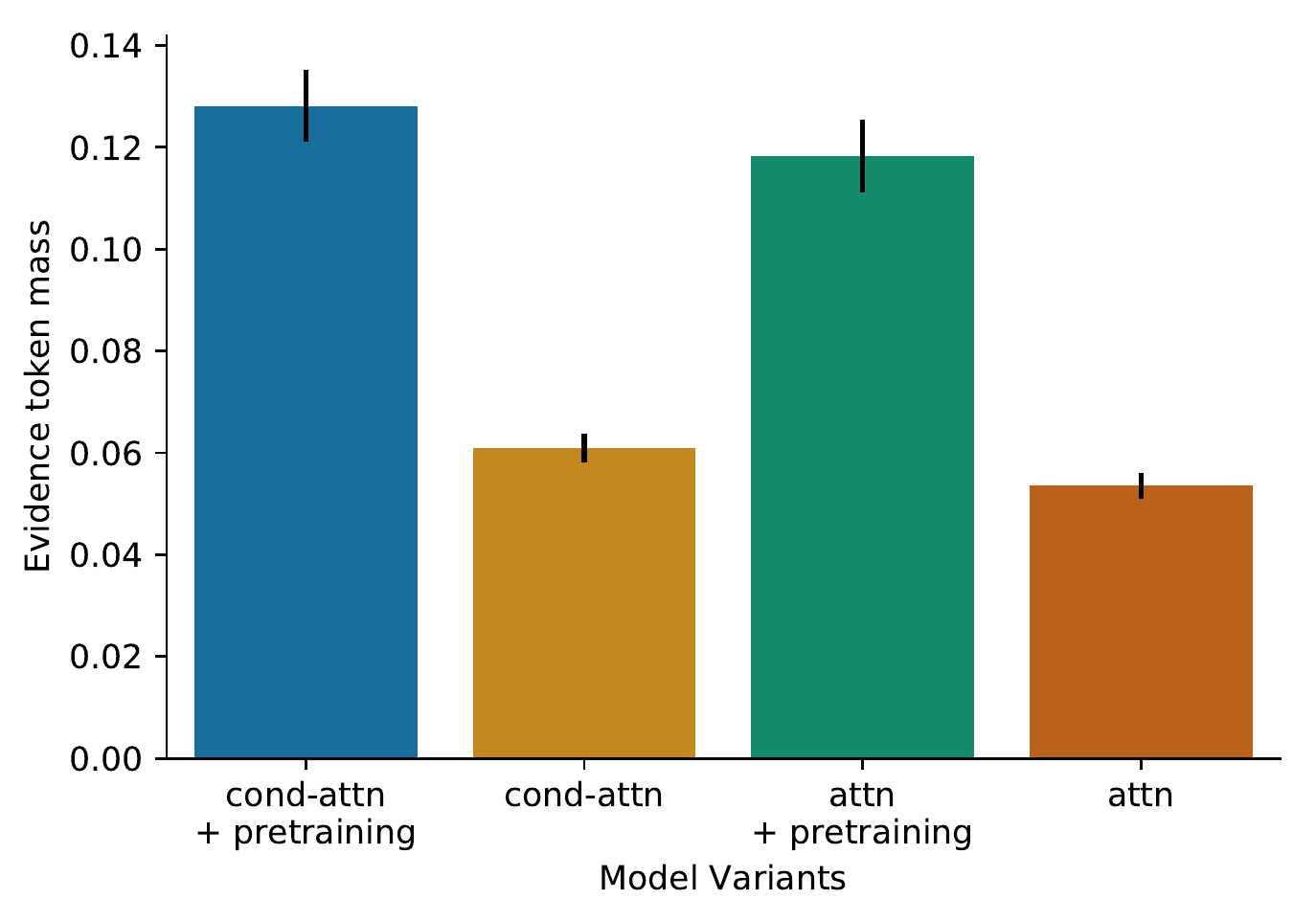} 
     \vspace{-.5em}
    \caption{Evidence token masses achieved by models on the validation set, after training.}
    \label{figure:evidence-token-mass}
     \vspace{-.5em}
  \end{figure}

Overall, despite conditioning and pretraining attention mechanisms, end-to-end models remain over 20 points behind the oracle variant (Table \ref{table:cheating}.
This suggests that the model is failing to sufficiently attend to the relevant evidence.
Figure \ref{figure:evidence-token-mass} supports this conjecture. 
This shows the total evidence mass realized by models on the validation set (after training for the final task of evidence inference). 
Even pretrained models assign $<$14$\%$ of total attention mass to actual evidence tokens. 

\begin{figure}
    \includegraphics[width=0.45\textwidth]{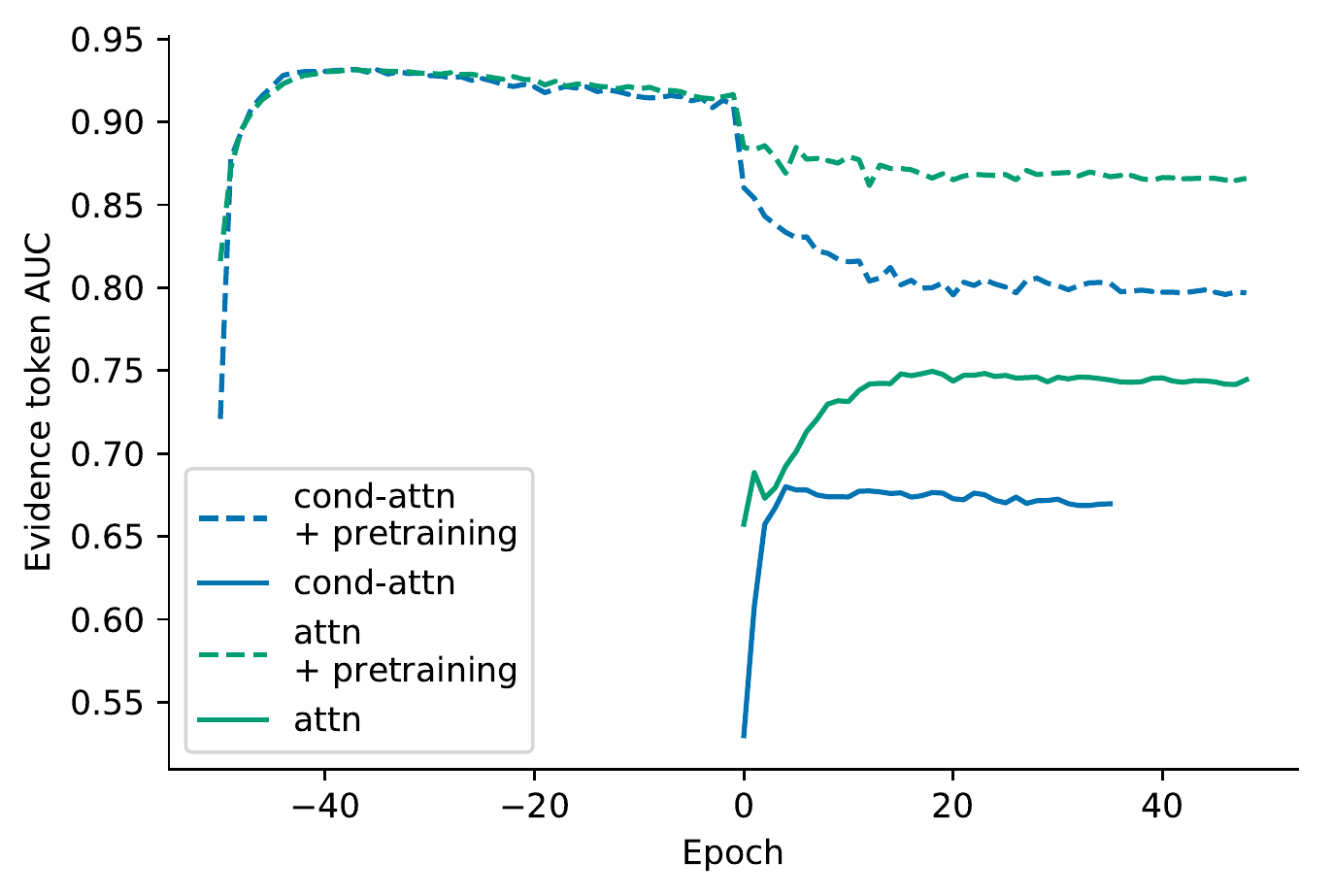}
    \vspace{-.5em}
    \caption{Validation evidence token AUCs during training. `pretraining' epochs are depicted as `negative' for the two explicitly supervised attention variants. Note that we use early stopping, so not all models run for the same number of epochs.}
    \label{figure:token-auc-over-epochs}
    \vspace{-.75em}
  \end{figure}

We also explore how token-level discriminative performance varies between pretrained and end-to-end variants (without explicit attention training), and how this changes as learning progresses.
Figure \ref{figure:token-auc-over-epochs} plots evidence token AUC over epochs (on the validation set) for attentive model variants (conditioned and unconditioned). 
We show curves for the case where we use explicit supervision (pretraining; dotted lines) and where relevance is learned only indirectly via the downstream evidence inference objective (no pretraining; solid lines).
Interestingly, evidence token AUC reaches maximum values during pretraining (shown as negative epochs) for supervised attention variants, and declines precipitously when the training objective transitions to the downstream task. 
This suggests a kind of catastrophic forgetting introduced due to shifting objectives. 

\section{Related Work} 
\label{section:related-work}

The proposed task is situated at the intersection of information extraction \cite{cardie1997empirical}, natural language inference \cite{bowman-EtAl:2015:EMNLP}, evidence mining \cite{rinott2015show} and question answering \cite{harabagiu2000experiments,hovy2000question}. 
However, our focus on inferring results from lengthy clinical trial reports pertaining to particular prompts constitutes a unique problem, as discussed in the Introduction. 

Prior systems have attempted to extract information from articles describing RCTs. For example, ExaCT \cite{Kiritchenko2010} attempts to extract variables describing clinical trials from articles, and ACRES \cite{summerscales-11} ingests extracts key variables from abstracts.
Blake and Lucic \shortcite{blake2015automatic,park2012identifying} considered the problem of automatically extracting interventions and outcomes in sentences that report direct comparisons. And Mihaila \emph{et al.} \shortcite{mihuailua2013biocause} have proposed annotating and extracting casual statements from biomedical literature.
Classifying the \emph{modality} of statements in scientific literature has also been investigated \cite{thompson2008categorising}; this relates to identifying evidence.

None of these prior efforts attempted to infer the \emph{findings} concerning the extracted interventions and outcomes, as we do here. 

\section{Conclusions and Future Work}
We have presented the task of inferring the polarity of comparative results reported in articles describing clinical trials with respect to interventions and outcomes of interest.
Such models would render the unstructured evidence currently buried in manuscripts actionable, in turn potentially informing evidence-based care.
In addition to the practical import of this problem, the task poses core NLP challenges related to processing lengthy, technical texts, and performing conditional inference over entities within these.

Our baseline results establish both the feasibility and difficulty of the task. 
Very simple baselines (e.g., rule-based methods) perform quite poorly, and modern neural architectures achieve the best results, currently.
When models are provided with reference evidence spans from an oracle, they achieve dramatically improved performance. 
This demonstrates that the key challenge concerns conditionally identifying relevant snippets to inform predictions; attention mechanisms would seem to provide a natural means of allowing the model to learn to focus, and we indeed found that (supervised) attention provides some predictive gains, but these are relatively modest. 

The gap between the model that directly consumes only relevant evidence snippets (Table \ref{table:cheating}) and the best performing end-to-end model is over 20 points in F1. 
Further, ignoring the article entirely (relying only on the prompt) degrades performance by only $\sim$5 points in F1, again suggesting that even the pretrained, conditioned attention variant is not making good use of the relevant evidence contained in articles. 
The evidence token mass metrics also support this: The best models we have proposed consistently place only $\sim$10-15\% of the attention mass on tokens actually marked as containing relevant evidence. 

We are simply not learning to attend well, even with explicit pretraining and conditioning.
This motivates a key future research direction: designing more sophisticated attention mechanisms that (conditionally) identify spans of evidence pertinent to a given prompt. We hope this corpus and task provides opportunity to pursue such models.


\section{Acknowledgements} 

This work was supported by NSF CAREER Award 1750978. 

We also acknowledge ITS at Northeastern for providing high performance computing resources that have supported this research. 

\bibliography{naaclhlt2019}
\bibliographystyle{acl_natbib}

\newpage 
\onecolumn
\appendix
\begin{center}
    {\large {\bf Appendix} }

\end{center}

\section{Description of Heuristics Baseline}

We describe the heuristics implemented in our rule-based baseline model. These can be broken into two stages: (1) finding the sentence that contains the answer, and (2) interpreting the sentence that possesses the evidence. The variant that consumes the evidence spans directly (thus "cheating") uses only rules defined for (2). In both cases we use a `points' based approach, where simple rules assign points to potential labels. 

As an attempt to identify evidence spans, we first split the article into sentences and tokenize these. Each sentence is then assigned a ranking based on the number of words in the outcome, intervention, and comparator that it contains; each map is associated with one point. The sentence with the highest number of points is then designated as being most likely to contain the evidence of interest.

Once this sentence is selected, we use simple checks to try and identify a reported $p$-value. In particular we search for the following three distinct forms: ``$p$ = $X$", ``$p$ $>$ $X$", and ``$p$ $<$ $X$"; we use a simple RegEx to find instances of these, ignoring whitespace. 

If we identify ``$p$ = $X$", we then attempt to identify the comparator and intervention from the prompt in the sentence, as the aim is to extract the $p$-value closest to the intervention or comparator (sentences may contain multiple $p$-values). If the $p$-value found is greater than 0.05, we add a point for a label of \emph{no significant difference}. If the $p$-value is smaller than 0.05, we add a point for a label of \emph{significant difference} (both significantly increases or significantly decreases). If neither the intervention nor the comparator is found in the selected sentence, then we look at all the $p$-values that it contains and sum points accordingly. 

In the case of ``$p$ > $X$", we add a point to the label of \emph{no significant difference}, as a $p$-value greater than some $X$ likely corresponds to a statistically non-significant result.

In the case of ``p $<$ 0.05", we add a point to a label of significant difference, as a p-value smaller than some X likely corresponds to varying effects for both the intervention and comparator with respect to the outcome. If there is a tie between points pointing to a significant difference (in either direction) and \emph{no significant difference}, then we return the latter, because it is the majority category.



Otherwise, we assume the result is significant, and next attempt to infer the reported direction of the effect. To do so, we count occurrences of synonyms of the word ``increase" in the sentence and compare that with the number of occurrences of synonyms of ``decrease". Synonyms are retrieved using WordNet via NLTK. If there are more synonyms of ``increase'', we return the label \emph{significantly increase}; if there are more synonyms of ``decrease", we return \emph{significantly decrease}. If the number of occurrences is equal, we return \emph{significantly increase}, as this designation is slightly more frequent than \emph{significantly decrease}.

\section{Annotation Costing Details} 

Generating each prompt cost an average of 99 cents. 
We thus paid about \$4.14 to complete prompt generation for each article. 
Doctors hired for prompt generation were paid an average of \$19.05 an hour. 

Prompts took an average of 2.54 minutes to complete.
Annotators (again MDs) were paid an average of \$13.88 dollars per hour, with each prompt costing 58 cents. 

Verification cost a mere 27 cents per prompt. 
Verifiers were paid, on average, \$16.25 an hour. 



\section{Additional Dataset Details}

Each prompt has at least two rows, one of which corresponds to the prompt generator's answer, and the other for an annotator's. These rows can be distinguished using the UserID column, in which prompt generator answers are marked with a value of `0', whereas annotator UserIDs are values greater than `0'. Each row also includes the verifier's response, which denotes the validity of that row's answer and rationale. We additionally include the offsets of where the rationale occurs in the text \footnote{Offset indices are based on the XML text post pre-processing}. These offsets are calculated through FuzzyWuzzy, due to the frequent differences in encoding between rationales and extracted XML text.

\section{Preprocessing Details}

To process the PubMed central XML documents, we iterate through by section, and parse subsequent sub-sections. Afterwards, we use an HTML parser library to remove all tags, as these might distract models. We removed all $<$p$>$ tags that remained due to malformed XML documents. 


\section{Attention (and Attention Pretraining) Variants}

On the validation set we explored several substantiations of attention and associatd objectives, in the case of pre-training. 
These include:

\begin{enumerate}
    \item {\bf Tokenwise}. This attention variant aims to maximize per-token evidence predictions independently, i.e., no softmax is imposed over attention activations, and so attention weights may range between 0 and 1. During the forward pass, however, we do normalize these weights prior to inducing the context vector.
    
    \item {\bf Balanced tokenwise attention}. This is the same as above, except that during training we construct samples composed of an equal number of evidence and non-evidence tokens (the latter far outnumbering the former, in general). 
    
    \item {\bf Evidence mass attention}. Here we attempt to directly optimize total quantity of attention mass placed on evidence tokens. 

\end{enumerate}

\noindent During pre-training, we run for some number of epochs over available marked evidence spans (we used a maximum of 50, with an early-stopping criterion using a patience of 10). This requires a metric to determine the `best' observed set of attention weights during pre-training (as measured on a nested validation set). For this we considered a few options for performance measuring criterion over tokens, including evidence mass and entropy. We ultimately settled on monitoring evidence token AUC as a proxy for attention performance over tokens. 

\section{Validation Results and Variances}

For completeness in Table \ref{val-results} we report means and ranges of all model variants explored on the validation dataset (taken over five runs). We include in these the attention variants explained above. 

For pretraining variants, trained on the observed evidence spans for 50 epochs prior to shifting objectives to the downstream evidence inference task. We then must select which set of weights, over these 50 epochs, to use for initialization following pretraining. We considered three metrics to score attentional components during pretraining: token-level AUC (overall discriminatory power of the network, with respect to evidence/not-evidence words); entropy (with the intuition that we seek peaky distributions); and evidence token mass (such that most attention is placed on evidence tokens). 

We settled on using the tokenwise pretrained variant along with token AUC as our pretraining criterion for the test experiments, but we report full results on the validation dataset in Table \ref{val-results}.

\begin{sidewaystable}
\begin{tabular}{lllll}
\hline
 Model                                                             & Precision            & Recall               & F1                   & Token AUC / Mass   \\
\hline
 + Attn.                                                           & 0.499 (0.470, 0.524) & 0.499 (0.461, 0.505) & 0.483 (0.460, 0.506) & 0.741 / 0.047      \\
 + Pretrain attn. [AUC] (Tokenwise attn.)                          & 0.514 (0.500, 0.529) & 0.514 (0.484, 0.506) & 0.497 (0.485, 0.508) & 0.882 / 0.131      \\
 + Pretrain attn. [Entropy] (Tokenwise attn)                       & 0.487 (0.446, 0.516) & 0.487 (0.432, 0.493) & 0.472 (0.433, 0.491) & 0.821 / 0.117      \\
 + Pretrain attn. [Evidence Mass] (Tokenwise attention)            & 0.501 (0.482, 0.525) & 0.501 (0.480, 0.512) & 0.493 (0.472, 0.513) & 0.861 / 0.128      \\
 + Pretrain attn. [AUC] (Tokenwise attn. balanced)                 & 0.507 (0.500, 0.523) & 0.507 (0.485, 0.516) & 0.500 (0.486, 0.517) & 0.803 / 0.062      \\
 + Pretrain attn. [Entropy] (Tokenwise attn. balanced)             & 0.489 (0.446, 0.499) & 0.489 (0.418, 0.488) & 0.470 (0.415, 0.491) & 0.704 / 0.046      \\
 + Pretrain attn. [Evidence Mass] (Tokenwise attn. balanced)       & 0.502 (0.499, 0.522) & 0.502 (0.465, 0.522) & 0.484 (0.464, 0.522) & 0.770 / 0.055      \\
 + Pretrain attn. [AUC] (Max evidence attn.)                       & 0.448 (0.440, 0.470) & 0.448 (0.416, 0.467) & 0.433 (0.411, 0.467) & 0.456 / 0.003      \\
 + Pretrain attn. [Entropy] (Max evidence attn.)                   & 0.458 (0.440, 0.471) & 0.458 (0.410, 0.472) & 0.436 (0.405, 0.471) & 0.413 / 0.000      \\
 + Pretrain attn. [Evidence Mass] (Max evidence attn.)             & 0.519 (0.514, 0.560) & 0.519 (0.491, 0.543) & 0.508 (0.492, 0.547) & 0.661 / 0.044      \\
 + Cond. attn.                                                     & 0.509 (0.510, 0.516) & 0.509 (0.478, 0.518) & 0.498 (0.477, 0.513) & 0.667 / 0.055      \\
 + Pretrain cond. attn. [AUC] (Tokenwise attn.)                    & 0.508 (0.500, 0.516) & 0.508 (0.488, 0.514) & 0.501 (0.488, 0.513) & 0.815 / 0.120      \\
 + Pretrain cond. attention [Entropy] (Tokenwise attn.)            & 0.494 (0.452, 0.525) & 0.494 (0.450, 0.521) & 0.482 (0.435, 0.517) & 0.813 / 0.171      \\
 + Pretrain cond. attn. [Evidence Mass] (Tokenwise attn.)          & 0.521 (0.509, 0.529) & 0.521 (0.501, 0.524) & 0.509 (0.499, 0.526) & 0.807 / 0.137      \\
 + Pretrain cond. attn. [AUC] (Tokenwise attn. balanced)           & 0.508 (0.506, 0.520) & 0.508 (0.489, 0.514) & 0.503 (0.489, 0.515) & 0.766 / 0.051      \\
 + Pretrain cond. attn. [Entropy] (Tokenwise attn. balanced)       & 0.466 (0.430, 0.500) & 0.466 (0.425, 0.477) & 0.454 (0.425, 0.479) & 0.620 / 0.032      \\
 + Pretrain cond. attn. [Evidence Mass] (Tokenwise attn. balanced) & 0.516 (0.520, 0.516) & 0.516 (0.487, 0.514) & 0.504 (0.487, 0.515) & 0.770 / 0.064      \\
 + Pretrain cond. attn. [AUC] (Max evidence attn.)                 & 0.495 (0.444, 0.506) & 0.495 (0.436, 0.499) & 0.482 (0.436, 0.500) & 0.617 / 0.042      \\
 + Pretrain cond. attn. [Entropy] (Max evidence attn.)             & 0.442 (0.428, 0.464) & 0.442 (0.414, 0.463) & 0.435 (0.413, 0.460) & 0.462 / 0.004      \\
 + Pretrain cond. attn. [Evidence Mass] (Max evidence attn.)       & 0.493 (0.474, 0.503) & 0.493 (0.474, 0.504) & 0.481 (0.470, 0.497) & 0.636 / 0.039      \\
\hline
\end{tabular}
\caption{Means and (min, max) values across five runs over the validation set for attention variants and associated (selection criteria). The latter is the metric used to pick the best set of weights after 50 pretraining epochs with which to seed the end-task training process.}
\label{val-results}
\end{sidewaystable}

On the test set, we only considered using AUC as the criterion.

\end{document}